\pdfoutput=1

\documentclass[11pt]{article}

\usepackage[preprint]{acl}

\usepackage{times}
\usepackage{latexsym}
\usepackage{lipsum}
\usepackage[T1]{fontenc}

\usepackage[utf8]{inputenc}

\usepackage{microtype}

\usepackage{inconsolata}

\usepackage{graphicx}
\usepackage{array}
\usepackage{booktabs}
\usepackage{tabularx}
\usepackage{amsmath}
\usepackage{tcolorbox}
\usepackage{colortbl}
\usepackage{xcolor}
\usepackage{amssymb}
\usepackage{multirow}
\usepackage{xspace}
\usepackage[english]{babel}
\usepackage{xurl}

\newif\ifcomment\commenttrue

\usepackage[a-1b]{pdfx}

\usepackage{framed}
\usepackage{mdwlist}
\usepackage{siunitx}
\usepackage{latexsym}
\usepackage{colortbl}
\usepackage{xcolor}
\usepackage{nicefrac}
\usepackage{booktabs}
\usepackage{fnpct}
\usepackage{amsfonts}
\usepackage[T1]{fontenc}
\usepackage{bold-extra}
\usepackage{amsmath}
\usepackage{amssymb}
\usepackage{bm}
\usepackage{graphicx}
\usepackage{mathtools}
\usepackage{microtype}
\usepackage{multirow}
\usepackage{multicol}
\usepackage{xpatch}
\usepackage{latexsym,comment}
\usepackage[normalem]{ulem}

\newcommand*{\missingreference}{{\Huge \colorbox{red}{?reference?}}}
\newcommand*{\missingcitation}{{\Huge \colorbox{red}{?citation?}}}

\makeatletter
\xpatchcmd{\@setref}{\bfseries}{\missingreference}{}{}
\def\@citex[#1]#2{\leavevmode
    \let\@citea\@empty
    \@cite{\@for\@citeb:=#2\do
        {\@citea\def\@citea{,\penalty\@m\ }%
            \edef\@citeb{\expandafter\@firstofone\@citeb\@empty}%
            \if@filesw\immediate\write\@auxout{\string\citation{\@citeb}}\fi
            \@ifundefined{b@\@citeb}{\hbox{\reset@font\missingcitation}%
                \G@refundefinedtrue
                \@latex@warning
                {Citation `\@citeb' on page \thepage \space undefined}}%
            {\@cite@ofmt{\csname b@\@citeb\endcsname}}}}{#1}}
\makeatother

\newcommand{\gem}[1]{\mbox{\textsc{gem}}}

\newcommand{\hidetext}[1]{}
\newcommand{\ignore}[1]{}

\ifcomment
    \newcommand{\pinaforecomment}[3]{\colorbox{#1}{\parbox{.8\linewidth}{#2: #3}}}

    \newcommand{\prtodo}[1]{\pinaforecomment{lightblue}{pr}{#1}}
    \newcommand{\prtodoi}[1]{\pinaforecomment{lightblue}{pr}{#1}}
\else
    \newcommand{\pinaforecomment}[3]{}
    \newcommand{\prtodo}[1]{}
    \newcommand{\prtodoi}[1]{}
\fi

\newcommand{\smallurl}[1]{ \begin{tiny}\url{#1}\end{tiny}}

\definecolor{lightblue}{HTML}{3cc7ea}
\definecolor{CUgold}{HTML}{CFB87C}
\definecolor{grey}{rgb}{0.95,0.95,0.95}
\definecolor{ceil}{rgb}{0.57, 0.63, 0.81}
\definecolor{UMDred}{HTML}{ed1c24}
\definecolor{UMDyellow}{HTML}{ffc20e}

\newcommand{\modelname}{\textsc{Navig}\xspace}
\newcommand{\dataname}{\textsc{NaviClues}\xspace}
\newcommand{\macname}{\textsc{Reasoner}\xspace}
\newcommand{\micname}{\textsc{Searcher}\xspace}
\newcommand{\guessname}{\textsc{Guesser}\xspace}
\newcommand{\geoloccap}{Geo-localization\xspace}
\newcommand{\geoloc}{geo-localization\xspace}

\title{\modelname: Natural Language-guided Analysis with Vision \\ Language Models for Image \geoloccap}

\author{Zheyuan Zhang$^{1}$\thanks{\quad Equal contribution.} \hspace{0.5cm} Runze Li$^{2}$\footnotemark[1] \hspace{0.5cm} Tasnim Kabir$^{3}$ \hspace{0.5cm} \textbf{Jordan Boyd-Graber}$^{3}$ \\
        $^{1}$Tsinghua University \hspace{0.5cm}
        $^{2}$Nanjing University \hspace{0.5cm}
        $^{3}$University of Maryland \\
        \texttt{zheyuan-22@mails.tsinghua.edu.cn} \hspace{0.5cm}
        \texttt{{jbg}@umiacs.umd.edu}
        }

\begin{document}

\maketitle 
\begin{abstract}
    Image \geoloc is the task of predicting the specific location of an image and requires complex reasoning across visual, geographical, and cultural contexts. While prior Vision Language Models (\textsc{vlm}s) have the best accuracy at this task, there is a dearth of high-quality datasets and models for analytical reasoning.  We first create \dataname, a high-quality dataset derived from GeoGuessr, a popular geography game, to supply examples of expert reasoning from language. Using this dataset, we present \modelname, a comprehensive image \geoloc framework integrating global and fine-grained image information. By reasoning with language, \modelname reduces the average distance error by 14\% compared to previous state-of-the-art models while requiring fewer than $1000$ training samples. Our dataset and code will be available at \url{https://github.com/SparrowZheyuan18/Navig/}.
\end{abstract}

\section{Introduction}

\begin{figure}[t]
  \includegraphics[width=\linewidth]{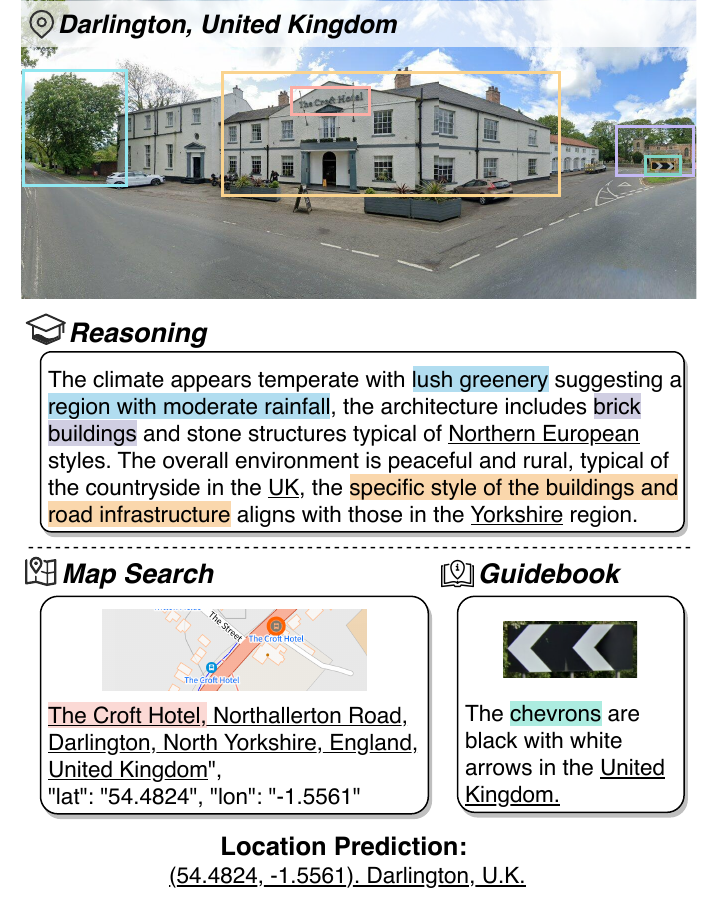} \caption{In
  image \geoloc, models need to find both cultural and
  geographical clues to infer correct locations. External tools like
  maps and guidebooks can also be helpful by providing extra
  knowledge.}  \label{fig:task}
\end{figure}

Image \geoloc---the task of predicting the location where an image
was taken~\cite{hays2008im2gps}---remains a challenging multimodal problem.
For example, to say Figure~\ref{fig:task} is a picture from
Darlington (in England) requires reading the name of the hotel to
determine possible candidates and excluding---for instance---the Croft hotel in Ontario based on architecture.
Directly predicting the exact location of an
image~\cite{weyand2016planet, haas2023learning, vivanco2024geoclip} is
difficult for computer vision models and requires extensive training on large
image-location datasets.  

In contrast, human experts infer locations by reasoning.
For example, in a GeoGuessr\footnote{\url{http://www.geoguessr.com}} game video, an expert player, \textit{zi8gzag}, explained how he identified a location in Korea: the presence of single yellow road lines and the language on the road signs suggest an Asian region; large spikes atop concrete poles narrow it down to Japan and Korea, and the black and yellow guardrails rule out Japan.
While recent research integrates textual knowledge~\cite{luo2022g} and explicit clues~\cite{zhang2024can, mendes2024granular, ligeoreasoner} with Vision Language Models (\textsc{vlm}s) to enhance accuracy, the reasoning in these models is often limited to a few words related to landmarks and does not provide a concrete analysis, as human experts would.

To date, these models' reasoning remains more superficial than humans' for two reasons: (1)~\textbf{Lack of high-quality reasoning datasets:} Existing geo-tagged datasets lack linguistic reasoning elements, while constructing a dataset that involves reasoning based on image details is resource-intensive. (2)~\textbf{Complexity of diverse information retrieval:} Images often contain rich details, such as road signs, texts, and building styles, requiring additional tools for accurate retrieval and interpretation.

To address these questions, we introduce \dataname, a detailed and high-quality reasoning dataset for image \geoloc, and \modelname, a framework that combines both visual analysis and external knowledge to perform analytical reasoning. Inspired by the popular game GeoGuessr, \dataname has over $2000$ instances from five experienced YouTubers, recording their process of analyzing image details to infer locations, which trains \textsc{vlm}s to generate reasoning that mimics professional human players. With tools like public maps and expert-written guidebooks, we design a pipeline that dives into fine-grained details and retrieves relevant information to further enhance accuracy. We evaluate \modelname against state-of-the-art models on two open benchmarks using five levels of prediction and ablate each component to investigate their contributions. \modelname outperforms previous state-of-the-art models by a 14\% reduction in average distance error while using less than $1000$ training samples. We further illustrate the reasoning of \modelname by providing examples of both successful and challenging cases. We release our dataset and framework to advance the use of reasoning in the field of image \geoloc.

\section{Collecting \dataname: Linking Places to Images}

This section explains how we process the reasoning of GeoGuessr players to construct \dataname. In addition, we analyze their reasoning and identify fifteen key clues humans use in \geoloc.

\subsection{Data Collection}
\label{sec:data_collection}

Despite previous datasets containing image--location pairs~\cite{hays2008im2gps, vo2017revisiting, astruc2024openstreetview} and reasoning insights from guidebooks~\cite{luo2022g, ligeoreasoner}, there is still a lack of datasets that capture the analytical reasoning process used to deduce locations from image details. To train \modelname to generate reasoning (Section~\ref{sec:macro}), we use the data from GeoGuessr, a popular game where players infer locations from street views, which preserve experts' knowledge and strategies for image \geoloc. We mine game data from ``play along'' videos of five popular YouTubers, along with transcripts of their reasoning during gameplay.

\textbf{Data Mining.} In a typical GeoGuessr game, there are multiple rounds of guessing the location from a new image. To segment the video transcript, we identify the timestamps of each round's result pages by using Qwen-VL~\cite{bai2023qwen} to match the buttons and extract the corresponding scores. For images, we retrieve images from Google StreetView (\textsc{gsv}) API based on the coordinates of each round, omitting any unavailable ones.\footnote{As \textsc{gsv} updates the images from time to time, some coordinates are deleted.} Following~\citet{haas2024pigeon}, we capture images from four different directions and combine them to create $360$-degree panoramic views (Table~\ref{tab:sft_data}), which contain the same details as in the games. For reasoning data, we split the transcripts by round timestamps. The raw dataset contains $2637$ images and respective locations.

\textbf{Data Processing.} To ensure data quality, we apply several processing steps: (1) we manually review and remove games where the visual content differs between the players' view and the \textsc{gsv} data, such as games that allow movement or feature satellite view; (2) we exclude games with poor reasoning quality, including those with a time limit of less than $30$ seconds, transcripts shorter than $100$ words, or incorrect answers (where the GeoGuessr Score is lower than $3400$, approximately $575$ km); (3) answer-guided reasoning~\cite{mendes2024granular} generates the step-by-step reasoning process from both the transcripts and images, guided by the correct locations and details mentioned by humans.

After processing, our final dataset, \textbf{\dataname}, consists of $1120$ images, each associated with a corresponding location (continent, country, and coordinates), reasoning, and scores. \dataname is geographically well-distributed (Table~\ref{fig:distribution}), covering a diverse set of images in various countries to reduce potential data biases. \dataname is useful for analyzing human \geoloc strategies and training models for related tasks. Prompts, examples, and other details are in the Appendix~\ref{apd:data}.

\textbf{Guidebook Data.} For \micname (Section~\ref{sec:micro}), we collect guidebook data from two popular GeoGuessr community sources,\footnote{\url{https://somerandomstuff1.wordpress.com/2019/02/08/geoguessr-the-top-tips-tricks-and-techniques} and \url{https://www.plonkit.net}} which include tips from top players of GeoGuessr. Following~\citet{luo2022g} and~\citet{ligeoreasoner}, each instance in our dataset contains a clue (\textit{\textit{e.g.}, The chevrons are black with white arrows in the U.K.}) paired with an image. The final dataset contains $6227$ image--clue pairs that have distinguishable geographical features, which could be further integrated into \geoloc frameworks as additional knowledge (Examples are in Appendix~\ref{apd:data}).

\subsection{Data Analysis}
\label{sec:data_analysis}

We further analyze the reasoning of human players and investigate common patterns among human experts in \geoloc tasks. We use spaCy~\cite{honnibal2015improved} for noun extraction, allowing us to explore the specific information these experts focus on. We manually filter out irrelevant words and phrases, retaining only content pertinent to geographic reasoning.

\begin{figure}[t!]
    \centering
    \includegraphics[width=0.9\linewidth]{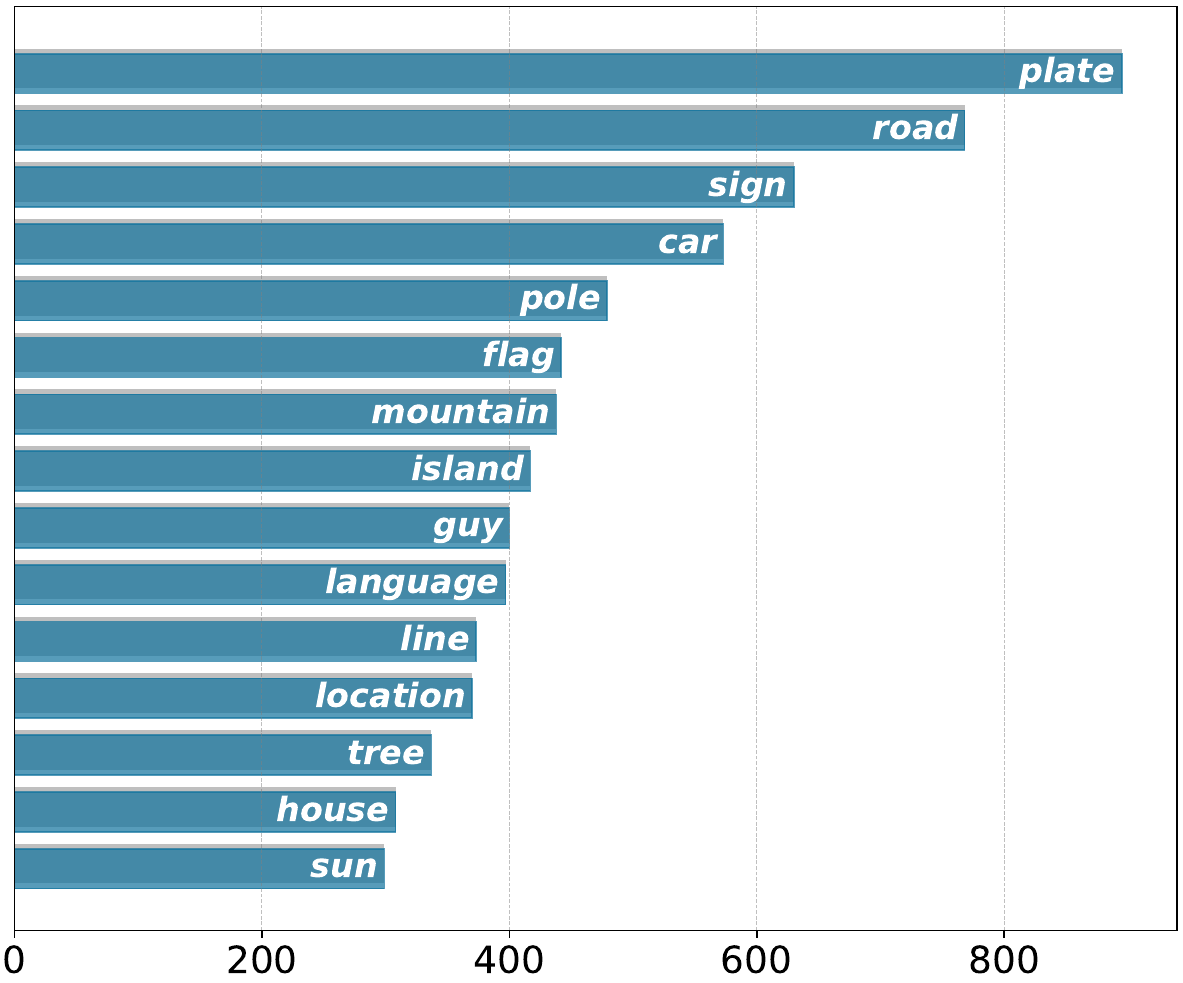} 
    \caption{Top clues in human reasoning. Humans identify \textit{roads}, \textit{cars}, \textit{poles}, and linguistic clues---specifically the languages on \textit{plates}, \textit{signs} and \textit{houses}.}
    \label{fig:clue-type}
\end{figure}

This process identifies fifteen core clues frequently mentioned by experts reflecting common analytical patterns and reasoning strategies (Figure~\ref{fig:clue-type}). The keywords cover cultural clues (\textit{e.g.}, language, flag, road, house) and natural geographical features (\textit{e.g.}, mountain, island, tree). This distribution conforms to categories in the guidebooks and further guides the implementation of our framework in Section~\ref{sec:method}.

\section{\modelname: Localizing Images with Reasoning and Tools}
\label{sec:method}

\begin{figure*}[t]
    \centering \includegraphics[width=\linewidth]{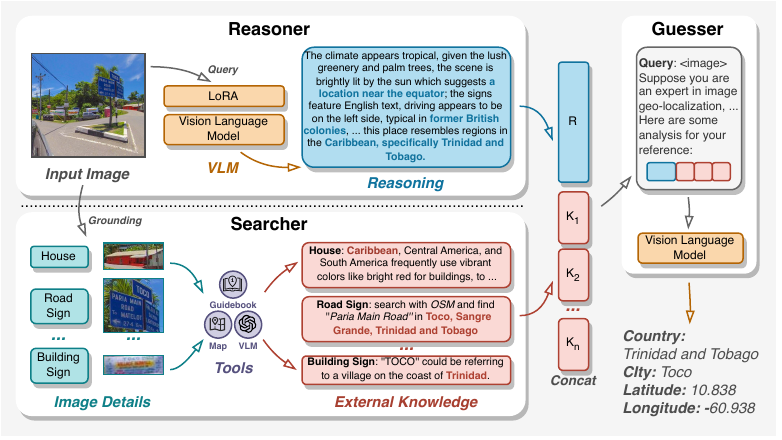}

\caption{The
    framework of \modelname comprises three main components:
    the \macname, which handles general reasoning; the \micname, which
    leverage external knowledge for detail-specific analysis, and
    the \guessname, which combines outputs from both analyzers to
    generate predictions.}  \label{fig:framework}
\end{figure*}

This section presents \modelname: image \geoloc with reasoning
about cultural and geographical clues and using
external tools. YouTube experts both reason with image elements (\textit{the driving is on the left}) and interpret image details with guidebooks or maps (\textit{the Paria Main Road is in Toco}) to locate an image. Based on this observation, \modelname (Figure~\ref{fig:framework}) has three components: (1)~\textit{\macname}, which focuses on generating a
reasoning process that analyzes the general information present in the
image (Section~\ref{sec:macro}); (2)~\textit{\micname}, which uses
additional knowledge sources and tools to explore the details
(Section~\ref{sec:micro}); and (3)~\textit{\guessname}, which takes
the concatenated outputs from both analyzers to determine the final
location, which can be configured to any granularity of locations
(Section~\ref{sec:guesser}).

\definecolor{climate}{HTML}{10739E}
\definecolor{vegetation}{HTML}{AE4132}
\definecolor{building}{HTML}{56517E}
\definecolor{infrastructure}{HTML}{36393D}

\begin{table}[t!]
    \centering
    \small
    \newcolumntype{Y}{>{\arraybackslash}X}
    \begin{tabularx}{\linewidth}{Y}
    \toprule 
        \textit{\textbf{Image}} \\
        \includegraphics[width=\linewidth]{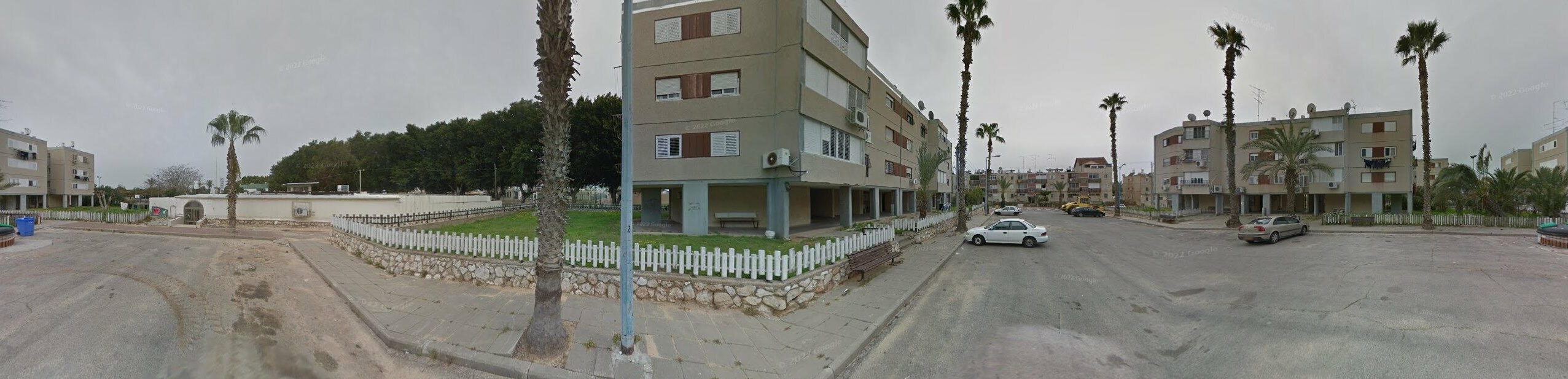} \\
        \midrule
        \textit{\textbf{Location}} \\
        Israel, Ashkelon. 31.66671, 34.59127 \\
        \midrule
        \textit{\textbf{Query}} \\
        <image> Given an image, craft a brief and cohesive reasoning path that deduces this location based on the visual clues present in the image. Using a tone of exploration ... \\
        \midrule
        \textit{\textbf{Response}} \\
        The scene reveals a blend of urban and natural features typical of \textcolor{climate}{\textbf{a Mediterranean climate}}. Notably, the presence of \textcolor{vegetation}{\textbf{palm trees}} suggests a \textcolor{climate}{\textbf{warm, temperate region}}. The architecture of the buildings, characterized by \textcolor{building}{\textbf{flat roofs and a modern, functional style}}, often indicates environments with minimal snowfall. ... \textcolor{infrastructure}{\textbf{the neatly paved roads and sidewalks with occasional dust}} align with urban areas in regions with dry summers. ... the collective visual attributes and environmental clues point to a location consistent with Israel's landscape and architectural style.\\
    \bottomrule
    \end{tabularx}
    \caption{Fine-tuning data format. The reasoning process leverages visual information in the images to deduce the correct location, such as \textcolor{climate}{\textbf{climate}}, \textcolor{vegetation}{\textbf{vegetation}}, \textcolor{building}{\textbf{building}}, and \textcolor{infrastructure}{\textbf{infrastructure}}. Complete prompts and response are in the Appendix.}
    \label{tab:sft_data}
\end{table}

\subsection{Training \textsc{vlm}s to \textsc{Reason} about Image Locations}
\label{sec:macro}

Recent \textsc{vlm}s can---sometimes---reason about the location of an
image.~\cite{ligeoreasoner}. However, the reasoning is limited to only
a few words and does not help localization~\cite{zhang2024can}.
To enhance \textsc{vlm}s to reason location-relevant
information in images, we create \dataname and fine-tune \textsc{vlm}s using it to build \macname. The
reasoning includes geographical information such as
climate, vegetation, building, and infrastructure (Table~\ref{tab:sft_data}). This approach
enables models to deduce locations from geographically pertinent
details, expanding the depth and applicability. 

After training, \macname can generate a rationale for images, where given an image~\( I \), the fine-tuned \textsc{vlm} produces a reasoning~\( R \). However, as the reasoning relies solely on \textsc{vlm}s constrained by their parameterized knowledge, it lacks the information to understand specific details. For instance, human experts can search maps for text on buildings or road signs and consult guidebooks to identify the house style of a particular country, which goes beyond the intrinsic knowledge within \textsc{vlm}s. To emulate this process, an additional module, \micname, integrates external tools, enabling more accurate interpretation of nuanced details.

\subsection{\textsc{Searching} Image Details}
\label{sec:micro}

The \micname module extracts fine-grained details from images to enhance the reasoning by integrating relevant knowledge. It crops the image, generates queries, and retrieves external knowledge.

\textbf{Grounding Image Details.} As highlighted in Section~\ref{sec:data_analysis}, human experts often concentrate on specific elements in images, such as signs, houses, and roads, which provide crucial location-based clues. A precise grounding model generates high-quality queries: given an image~\( I \) and a predefined set of elements~\( E = \{e_1, e_2, \dots, e_n\} \), \micname uses GroundingDino~\cite{liu2023grounding} to crop the image according to \(E\). Since each image may contain multiple instances of an element, the cropped images is defined as \( C = \{ c_{i,j} \mid e_i \in E, j \in [1, m_i] \} \), where \( m_i \) is the count of element~\( e_i \) in \( I \). Specifically, we select \textit{road sign}, \textit{building sign}, and \textit{house} from Figure~\ref{fig:clue-type} as elements, which align well with GroundingDino, since alternatives could yield overly large figures or uninformative results. Each cropped image \( c_{i,j} \) is a query for specific tools. Additionally, if \( c_{i,j} \) is a \textit{sign} that contains text, text-based queries are generated with Optical Character Recognition (\textsc{ocr}) from Qwen2-VL~\cite{Qwen2VL}. Therefore, the query set~\( Q \) is:

\vspace{-5pt}
\[
Q = \bigcup_{i,j} \{ c_{i,j}, \text{OCR}(c_{i,j}) \text{ if } c_{i,j} \in \text{signs} \}
\]

\textbf{Tools.} The query set~\( Q \) is then fed into a Tool Set~\( T \), which retrieves relevant knowledge. We use three tools for information retrieval: 
(1) The \textit{GeoGuessr Guidebook} contains rich information for locating images (Section~\ref{sec:data_collection}). Following prior research~\cite{luo2022g, zhou2024img2loc}, we frame \textit{Guidebook} using as a Retrieval-Augmented Generation problem. Given an input image (\textit{e.g.}, a house as in Figure~\ref{fig:framework}), we retrieve the most similar images. 
(2) \textit{Map.} The map is a critical tool in image \geoloc: text in images (\textit{e.g.}, a name on a sign) can pinpoint a location. We use OpenStreetMap\footnote{\url{https://www.openstreetmap.org/}} for location retrieval, providing the top three search results, with the place name and multi-level location details.
(3) \textit{\textsc{vlm}.} We use an additional \textsc{vlm} as a tool by prompting it to identify details that might be overlooked in the \macname. The \textsc{vlm} generates descriptions for details to narrow down potential locations (Figure~\ref{fig:framework}). Each tool~\( t \) in the Tool Set~\( T \) contributes to the retrieval of additional knowledge~\( K \): 

\vspace{-5pt}
\[
K = \bigcup_{t \in T} t(Q)
\]
Further implementation details in the Appendix~\ref{apd:implementations}.

\subsection{\textsc{Guessing} the Final Location}
\label{sec:guesser}
The \guessname uses all prior information to generate the final prediction. It concatenates the reasoning $R$ from the \macname with the external knowledge $K$ retrieved by the \micname, forms them into a prompt template $p$ along with the image $i$, and makes the location prediction with a \textsc{vlm}:

\vspace{-5pt}
\[
\hat{y}_{\text{loc}} = \mathrm{\textsc{vlm}}_{p} \big( I, \mathrm{concat}(R, K) \big)
\]
where $\hat{y}_{\text{loc}}$ is the model's generated location. The prompt $p$ is configurable to flexibly adjust to specific output formats, such as various location levels (\textit{e.g.}, country, city, and coordinates).

\section{How Well Does \modelname Reason Image Locations?}

We compare \modelname against prior state-of-the-art image \geoloc models and other baseline approaches (Section~\ref{sec:main_experiments}), ablate each module to evaluate their contributions (Section~\ref{sec:ablation_study}), and provide qualitative examples to highlight successful and challenging cases (Section~\ref{sec:qualitative_analysis}).

\subsection{Experimental Setup}

\textbf{Implementation.} We use three open-source models in \modelname: \mbox{MiniCPM-V~\cite{yao2024minicpm}}, LLaVA~\cite{liu2024llavanext}, and \mbox{Qwen2-VL}~\cite{Qwen2VL}. These models serve as \textsc{vlm}s for \macname, \micname, and the \guessname components within the \modelname framework. (1)~For \macname, Low-Rank Adaptation (LoRA)~\cite{hu2021lora} fine-tunes models using \dataname. We use \mbox{\textit{minicpm-v-2.6}}, \mbox{\textit{llava-1.6-vicuna-7b}}, and \mbox{\textit{qwen2-vl-7b}} for their advanced performance and mid-range size, which align with our cost constraints.
(2)~For \micname, we select the top three cropped clues as the basis for generation (\textit{e.g.}, if multiple houses are cropped, only will the three with the highest similarity be analyzed). \textsc{clip}~\cite{radford2021learning} encodes guidebook images and query images, retrieving guidebook data by the Euclidean distance~\(d\) between image embeddings (\textsc{faiss}~\cite{johnson2019billion}), returning associated text if \(d\) is below a threshold~$d_t$ (set to 30). 
(3)~We prompt the \guessname to predict locations at the coordinates level. Training hyperparameters, model configurations, and prompts are in Appendix~\ref{apd:implementations}.

\begin{table*}[t]
\centering
\small
\begin{tabular}{l r r r r r r r}
\toprule
\textbf{Model} & \multicolumn{1}{c}{\begin{tabular}[c]{@{}c@{}}\textbf{Continent}\\ $2{,}500$ km\end{tabular}} & 
\multicolumn{1}{c}{\begin{tabular}[c]{@{}c@{}}\textbf{Country}\\ $750$ km\end{tabular}} & 
\multicolumn{1}{c}{\begin{tabular}[c]{@{}c@{}}\textbf{Region}\\ $200$ km\end{tabular}} & 
\multicolumn{1}{c}{\begin{tabular}[c]{@{}c@{}}\textbf{City}\\ $25$ km\end{tabular}} & 
\multicolumn{1}{c}{\begin{tabular}[c]{@{}c@{}}\textbf{Street}\\ $1$ km\end{tabular}} & 
\multicolumn{1}{c}{\textbf{\begin{tabular}[c]{@{}c@{}}Distance$\downarrow$\end{tabular}}} & 
\multicolumn{1}{c}{\textbf{\begin{tabular}[c]{@{}c@{}}Score$\uparrow$\end{tabular}}}\\
\midrule
 $G^3$             & 50.9 & 14.6 &  2.3 &  0.1 & 0.0 & 4,341 & 1,304 \\
GeoCLIP           & 78.2 & 46.5 & 17.1 &  3.5 & 0.4 & 2,099 & 2,613 \\
StreetCLIP        & 79.4 & 43.4 & 13.4 &  1.7 & 0.3 & 2,060 & 2,543 \\
\midrule
MiniCPM-V         & 27.1 &  15.9 &  6.7 &  1.6 & 0.1 & 7,320 &   909 \\
 LLaVA             & 43.9 & 23.1 &  7.0 & 1.2 & 0.0 & 5,096 & 1,418 \\
Qwen2-VL          & 89.4 & 66.7 & 31.8 &  6.1 & 0.1 & 1,124 & 3,344 \\
\midrule
 \textit{\textbf{\modelname}} &&&&&&& \\
~~ - MiniCPM-V    & 71.5 & 44.1 & 16.9 &  3.5 & 0.3 & 2,956 & 2,413 \\
 ~~ - LLaVA        & 74.7  & 39.4 & 12.0 &  1.9 & 0.3 & 2,243 & 2,354 \\
~~ - Qwen2-VL     & \textbf{91.1} & \textbf{66.9} & \textbf{31.9} &  \textbf{6.7} & \textbf{0.7} & \textbf{965} & \textbf{3,389} \\
\bottomrule
\end{tabular}
\caption{
Accuracy and scores on GWS5k. The data from Continent to Street represents the accuracy (\%) at each level. The three sections are \geoloc models, \textsc{vlm}s, and \modelname. \textbf{Bold} font indicates the best performance. \modelname (\mbox{Qwen2-VL}) achieves the highest accuracy across all metrics. 
}
\label{tab:main_results}
\end{table*}

\textbf{Baselines.}
We compare \modelname with two baselines: (1) \textit{\geoloccap Models}: we select top-performing open-source models from prior research in image \geoloc: $G^3$~\cite{luo2022g}, GeoCLIP~\cite{vivanco2024geoclip}, and StreetCLIP~\cite{haas2023learning}. (2) \textit{Vision Language Models}: we select vanilla MiniCPM-V, LLaVA, Qwen2-VL as baselines, consistent with the backbone models used in \modelname. The prompts for these \textsc{vlm} baselines are identical to those in \modelname but lack analyses. We do not include commercial closed-source models (as discussed in Limitations).

\textbf{Dataset and Metrics.}
Following previous work~\cite{hays2008im2gps, astruc2024openstreetview, haas2024pigeon}, we evaluate our framework on two public datasets, including GWS5K sampled from GWS15K~\cite{clark2023we} due to cost constraints, and Im2GPS3k~\cite{hays2008im2gps}. We first computed the haversine distance between predicted and ground truth coordinates. For models limited to city level outputs, we use the coordinates of the predicted city as their predictions. Next, we evaluated the prediction accuracy---the percentage of guesses that fall within a distance threshold from the correct location---at five geographic levels: Street ($1$ km), City ($25$ km), Region ($200$ km), Country ($750$ km), and Continent ($2{,}500$ km). In addition, we calculated the average error distance and GeoGuessr Score, a metric from the original GeoGuessr game that quantifies guess accuracy, with a scoring range of $0$ to $5000$. Details about metric computation are in Appendix~\ref{apd:metrics}.

\subsection{Main Experiments}
\label{sec:main_experiments}

\begin{table}[t!]
\centering
\small
\begin{tabular}{l c c c }
\toprule
\multirow{2}{*}{\textbf{Model}} & \multicolumn{3}{c}{\textbf{\textsc{rouge} F1}} \\
 & \textbf{R1} & \textbf{R2} & \textbf{RL} \\
\midrule
 \textbf{\macname}~(MiniCPM-V)    & 51.0 & \textbf{14.8} & \textbf{24.6} \\
MiniCPM-V         & 46.4 & 12.6 & 22.1 \\
 \textbf{\macname}~(LLaVA)  & 49.8 & 13.9 & 24.0 \\
LLaVA       & 44.7 & 10.8 & 21.8 \\
 \textbf{\macname}~(Qwen2-VL)   & \textbf{51.4} & 14.6 & 24.3 \\
Qwen2-VL        & 45.2 & 12.3 & 22.1 \\
\bottomrule
\end{tabular}
\caption{\textsc{rouge} F1 scores for reasoning generated by models and humans (\%). \macname models reason more similarly to humans.}
\label{tab:rouge_f1_scores}
\end{table}

\textbf{Accuracy.} We compare \modelname with state-of-the-art image \geoloc models and Vision Language Models (GWS5k results in Table~\ref{tab:main_results}). (1)~\modelname(Qwen2\mbox{-}VL) has the highest accuracy across all metrics, beating specialized \geoloc models trained on domain-specific datasets, despite its relatively compact size of only 7 billion parameters. (2)~All \textsc{vlm}s generate effective analytical reasoning trained with only around $1{,}000$ samples and beat their vanilla models.
These findings underscore the quality of training data and the efficacy of \modelname. Similar results on Im2GPS3k are in Appendix~\ref{apd:supplementary_exps}.

\textbf{Reasoning.} We evaluate the linguistic reasoning quality generated by the model on a reserved test set of 50 games. To measure the alignment between model and human reasoning, we compute their \textsc{rouge} scores~\cite{lin2004rouge} which illustrate whether the model simulates human reasoning. \macname achieves higher \textsc{rouge} scores across all models and metrics after training (Table~\ref{tab:rouge_f1_scores}). 

We apply GPT-4o to label the granularity and accuracy (measured by exact match) of the reasoning by \modelname(Qwen2-VL) on GWS5k (Table~\ref{tab:reasoning}). \macname predicts country with an accuracy of 79.6\%, while it's challenging (3.0\%) when it makes finer-grained predictions (\textit{e.g.}, city, town, or street), as these predictions require additional information. For example, \macname outputs ``the combination of these elements point towards a city like Chaco, Argentina'' while the correct answer is ``Trelew, Argentina.'' This indicates the importance of \micname for precise predictions.

\begin{table}[t!]
\centering
\small
\begin{tabular}{l r r }
\toprule
\textbf{Level} & \textbf{Frequency} & \textbf{Accuracy} \\
\midrule
Country & 100.0 & 79.6 \\
Others & 50.7 & 3.0 \\
\bottomrule
\end{tabular}
\caption{Frequency and accuracy (exact match) of \macname. ``Others'' indicates more detailed predictions, which are challenging.}
\label{tab:reasoning}
\end{table}

\textbf{Comparison with Humans.} 
We also compare \modelname's performance against human players in fifty randomly sampled GeoGuessr games (collected in Appendix~\ref{apd:data}), focusing on common metrics for country, city, and street level predictions. Filtered human player data have time and access to additional knowledge resources (\textit{e.g.}, maps and guidebooks) for a fair comparison with \modelname. \modelname outperforms humans in overall scores (Table~\ref{tab:human_comparison}), although humans excel at finer-grained predictions by iteratively cross-referencing maps and comparing terrain and features within the game. This highlights a future direction to use non-textual features to refine map-based searches and enhance street-level accuracy of models.

\begin{table}[t!]
\centering
\small
\begin{tabular}{l r r r r}
\toprule
\textbf{Model} & \textbf{Country} & \textbf{City} & \textbf{Street} & \textbf{Score$\uparrow$} \\
\midrule
 \textit{\textbf{\modelname}} &&&& \\
~~ - MiniCPM-V & 56.0 & 18.0 & 0.0 & 2,863 \\
~~ - LLaVA & 48.0 & 14.0 & 0.0 & 2,690 \\
~~ - Qwen2-VL & \textbf{86.0} & 32.0 & 4.0 & \textbf{4,202}\\
\midrule
\textbf{Human Players} & 76.0 & \textbf{48.0} & \textbf{42.0} & 3,757\\
\bottomrule
\end{tabular}
\caption{Performance between humans and \modelname. The data from City to Street represents accuracy (\%). Our best model beats humans with a higher overall score but still struggles to achieve fine-grained accuracy.}
\label{tab:human_comparison}
\end{table}

\subsection{Ablation Study}
\label{sec:ablation_study}

To illustrate the contributions of each component in \modelname, we ablate the reasoning training, the impact of \macname, and \micname. Table~\ref{tab:ablation_results} presents the three \textsc{vlm}s' accuracy on GWS5k. In this setup, \modelname represents our framework, ``w/o training'' denotes results with the same prompt but without training on \dataname, ``w/o Macro'' and ``w/o Micro'' refer to the results without the \macname and \micname modules, respectively. 

\begin{table}[t!]
\centering
\small
\begin{tabular}{l r r r}
\toprule
\textbf{Model} & \textbf{Country} & \textbf{City} & \textbf{Street} \\
\midrule
 \textbf{\modelname}~(MiniCPM-V) & \textbf{44.1} &  \textbf{3.5} &  \textbf{0.3} \\
~~ - w/o training             &- 3.3 & - 0.4 & - 0.2 \\
 ~~ - w/o \macname                &- 10.2 & - 0.7 & - 0.0 \\
~~ - w/o \micname                & - 0.3 & - 0.3 & - 0.2 \\
 ~~ - MiniCPM-V                &- 14.9 & - 0.5 & - 0.2 \\
\midrule
 \textbf{\modelname}~(LLaVA) & 39.4 & \textbf{1.9} &  \textbf{0.3} \\
~~ - w/o training             &- 25.8 & - 1.2 & - 0.3 \\
 ~~ - w/o \macname                &- 20.2 & - 0.8 & - 0.0 \\
~~ - w/o \micname                & \textbf{+ 0.4} & - 0.2 & - 0.2 \\
 ~~ - LLaVA                    &- 16.3 & - 0.7 & - 0.3 \\
\midrule
 \textbf{\modelname}~(Qwen2-VL) & 66.9 &  \textbf{6.7} &  \textbf{0.7} \\
~~ - w/o training             & - 6.0 & - 0.9 & - 0.5 \\
 ~~ - w/o \macname                & - 4.0 & - 0.6 & - 0.2 \\
~~ - w/o \micname                & \textbf{+ 0.1} & - 0.9 & - 0.5 \\
 ~~ - Qwen2-VL                 & - 0.2 & - 0.6 & - 0.6 \\
\bottomrule
\end{tabular}
\caption{
Ablation results of \modelname on the GWS5k dataset. Each component contributes to model accuracy, with their removal leading to notable declines across Country, City, and Street levels.
}
\label{tab:ablation_results}
\end{table}

\textbf{Results.} (1)~Each module contributes to improving the model's accuracy. (2)~Surprisingly, when the model is prompted to zero-shot generate reasoning, it can be misleading and decrease final accuracy. This highlights the necessity of training with \dataname. (3)~\macname plays a critical role in coarse-grained localization, with improvements at the country level and decreases without it, as the reasoning in \dataname is limited to the country and city level. (4)~\micname substantially enhances fine-grained reasoning. Precise street-level localization on the GWS dataset is challenging, but the \micname narrows the scope within 1 km for images containing textual information using map searches (Table~\ref{tab:main_results}). Results on Im2GPS3k are in Appendix~\ref{apd:supplementary_exps}, which is consistent with GWS.

\subsection{Qualitative Analysis}
\label{sec:qualitative_analysis}

\begin{figure*}[t!]
    \centering
    \includegraphics[width=0.827\linewidth]{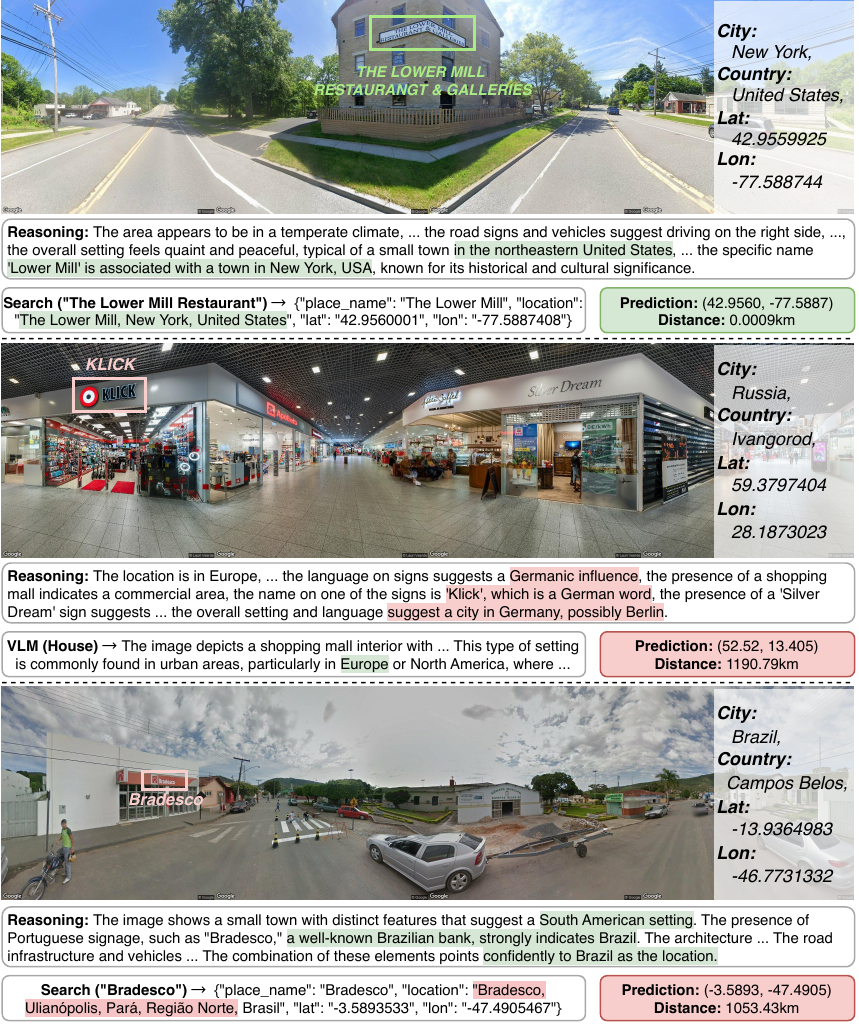} 
    \caption{\textbf{Top:} The model uses visual details and OpenStreetMap to accurately determine the location. \textbf{Middle:} The model is misled by linguistic elements---the shop name, resulting in an incorrect inference. \textbf{Bottom:} The model found a namesake when using OpenStreetMap.}
    \label{fig:cases}
\end{figure*}

This section examines how the analytical reasoning derived from images contributes to \modelname’s inference process. \modelname closely examines details within the image (\textit{e.g.}, climate, orientation, and ``Lower Mill'' in Figure~\ref{fig:cases} (top)) to determine the location. This detailed reasoning narrows down the possible range, while integration with OpenStreetMap data further aids the model in finding the restaurant, with an error distance of under $1$ meter, improving its estimate by $144$ km.

However, image elements can also mislead the model. In Figure~\ref{fig:cases} (middle), the model fixates on a shop name in the image, ``KLICK'', which can be interpreted as a German word. This leads the reasoning process astray, resulting in an incorrect localization. OpenStreetMap can also lead to false predictions when there are places with the same name, such as ``Bradesco'', a well-known Brazilian bank (Figure~\ref{fig:cases}, bottom). The reasoning makes image \geoloc models more interpretable by revealing how image elements influence decisions.

\section{Related Work}

\textbf{Image \geoloc.}
Image \geoloc falls into three
methods: (1) \textit{Retrieval-based methods}
retrieves the most similar images~\cite{hays2008im2gps, zhu2023r2former}. Various
retrievers~\cite{vo2017revisiting, pramanick2022world,
haas2023learning} and gallery types~\cite{vivanco2024geoclip} have
been proposed.
(2) \textit{Classification-based
methods} divide geographical maps into distinct classes
and train models to classify
the images into these categories with different model structures~\cite{radford2021learning,
wu2022im2city} and map division strategies~\cite{weyand2016planet,
theiner2022interpretable, haas2024pigeon}. 
(3) \textit{Generation-based methods} use visual understanding and generation in Vision
Language Models (\textsc{vlm}s) to directly generate the location or
coordinates for \geoloc. Aligning visual content with rich
text descriptions and reasoning~\cite{jia2024g3, ligeoreasoner,
zhang2024can} and incorporating external knowledge through
Retrieval-Augmented Generation~\cite{luo2022g, zhou2024img2loc} are state-of-the-art. However, challenges
persist in effectively using \textsc{vlm}s, including limited reasoning
data and relying on constrained knowledge
sources.

\textbf{Visual Reasoning.}
Unlike previous methods, we treat \geoloc as a complex visual reasoning task that deduces the correct location with language, requiring both visual
understanding and reasoning capabilities~\cite{hudson2019gqa,
gupta2023visual}.
As \textsc{vlm}s have demonstrated exceptional accuracy in visual
reasoning tasks~\cite{alayrac2022flamingo, lu2024chameleon}, methods enhancing the visual reasoning of \textsc{vlm}s in specific tasks include:
(1) \textit{High-quality reasoning data}, which researchers
have shown to be particularly effective in improving the performance
of \textsc{vlm}s~\cite{du2023makes, chen2023shikra}; (2) \textit{Vision
grounding}, which enables models to ground in the details of the image
and perform step-by-step reasoning~\cite{qi2024cogcom, wu2023textit,
zhang2024can}; and (3) \textit{Tool using and retrieval-augmented generation},
which aid the model by leveraging tools to retrieve additional knowledge~\cite{yang2023mm, marino2021krisp, chen2022lako} and reduce hallucinations. 
We integrate these insights to improve \modelname.

\section{Conclusion}

We introduce a reasoning dataset \dataname and a novel framework \modelname with detailed visual reasoning and knowledge retrieval for image \geoloc. Evaluated with relatively small backbone models, \modelname is highly accurate by using the reasoning chains of human experts, providing interpretable, verifiable evidence that a downstream user can use to validate its predictions. Future work could include detailed annotations to images more than street views, expanding more tools to improve finer-grained predictions, and using interpretable reasoning to assist \geoloc applications.

\section*{Limitations}

\subsection*{Dataset} 

(1) \textit{Limited data size.} In this work, we utilize data from human players in the GeoGuessr game to train Vision Language Models for performing geographic reasoning on images. The copyright and usage rights of the images are subject to that of Google Street View. However, the size of \dataname is limited due to the scarcity of available data on YouTube and the data noise. 

(2) \textit{Panoramic street view images.} To simulate the perspective of players in the GeoGuessr game, we use stitched panoramic images as the input to the model. Furthermore, nearly all images in the data from GeoGuessr are street views, despite our efforts to ensure a geographically balanced distribution of data across countries. This limits its distribution, as there's more weather, street, car, and vegetation information in street views than in other images. Models trained with \dataname might be weak at images with less street-level information. 

(3) \textit{Future work} could consider expanding the training dataset by incorporating images of different sizes and types, including more detailed annotations to create dataset s more than street views, to further enhance the performance of image \geoloc tasks with better reasoning.

\subsection*{Models} 

(1) \textit{Limited model sizes.} Due to cost constraints, we are unable to train larger models and conduct our experiments using top-performing, medium-sized open-source models (around 7B parameters). While this choice may result in performance that is not as competitive as larger models, it ensures a practical balance between computational feasibility and model efficacy. We also refrain from using closed-source models, as their lack of transparency regarding training data and inability to be trained on \dataname make them unsuitable for fair comparison. 

(2) \textit{Limited tool sets.} We evaluated only a limited set of tools and grounding words in \micname. Identifying more geographic features such as cars, road markings, and poles would require more precise recognition methods and more sophisticated model designs, which could potentially improve performance. 

(3) \textit{Complexity of subsystems.} We employ a pipeline approach to construct our model, aiming to maximize the performance of each component at every stage. However, this process introduces knowledge from different resources, which might conflict with each other. Currently, we implement a Guesser to handle the potential conflict and show the contribution of each ablated subsystems. We also examine the reasoning from \macname to show the necessity of \micname.

(4) \textit{Future works} can focus on including larger backbone models to further improve the performance, adding more tools, and conduct end-to-end training to better integrate the information, or add another fact-checking module to better discern information.

\section*{Ethical Considerations}

\subsection*{Data Collection}

In this work, we use the data from GeoGuessr players on YouTube to train our models. We carefully process the data and remove the personal information of the players, using all data for academic and non-commercial purposes, and giving appropriate credit to them in this paper. We make sure the use of our data is acceptable under YouTube's copyright policies and the Fair Use guidelines. 

\subsection*{Model Usage}

While the task of image \geoloc has the potential to enable innovative applications in fields such as navigation and tourism, the misuse of these models could also lead to risks such as privacy breaches and surveillance. In our work, we ensured that all training and testing data came from publicly available sources, with no involvement of private or personal images or location data. Currently, as shown in our experiments, these models have not yet reached a level of precision to accurately predict coordinates-level locations. For the future development of this field, it is crucial for researchers to ensure that these models are used within appropriate boundaries to prevent the leakage of private information. 

\bibliography{custom}

\clearpage

\appendix

\section{Implementation Details}
\label{apd:implementations}

\subsection{Training Parameters}

We trained the \macname on Nvidia RTX 6000 Ada (48G), with CUDA 12.4, Transformers 4.45.1, and Pytorch 2.1.2.

\begin{table}[htbp]
\centering
\small
\begin{tabular}{l c}
\toprule
\textbf{Parameters} & \textbf{Value} \\
\midrule
Max Length & 2048 \\
LoRA Rank & 8 \\
LoRA Alpha & 32 \\
Optimizer & AdamW \\
Adam Beta1 & 0.9 \\
Adam Beta2 & 0.95 \\
Learning Rate & 1e-4 \\
Warmup Ratio & 0.05 \\
LR Scheduler Type & cosine \\
Batch Size & 1 \\
Weight Decay & 0.1 \\
\bottomrule
\end{tabular}
\caption{
Training parameters for \macname.
}
\label{tab:training_parameters}
\end{table}

\subsection{Other Parameters.}

For reproducibility, we also provide the parameters used in other modules and \textsc{vlm}s within our framework.

\textbf{GroundingDino.} We utilize GroundingDino to crop detailed information from the images, such as signs and houses. We observe variation in the features of images across different datasets. For instance, the GWS5k dataset focuses on street scenes, and other datasets contain considerable noise (\textit{e.g.}, animals). Consequently, to reduce noise that could potentially affect model performance, we empirically set the thresholds as follows: \textit{Box-Threshold} = 0.5 and \textit{Text-Threshold} = 0.5 for GWS5K, and \textit{Box-Threshold} = 0.8 and \textit{Text-Threshold} = 0.6 for Im2GPS3k.

\textbf{Retrieval-Augmented Generation.} We employ CLIP as the image encoder for guidebook clues, using ViT-B-32 as the vision encoder. The guidebook database is deployed with \textsc{faiss}, and similarity is calculated using Euclidean Distance. The number of most relevant retrieved images, $k$, is set to 3, with a similarity threshold of 30.

\textbf{OpenStreetMap.} We use the \textit{Nominatim Search API} to process map searches, which takes text queries, and return the most relevant results along with the place name, address, and coordinates.

\textbf{Vision Language Models.} We use Vision-Language Models in our framework for reasoning and location inference. The three models are \mbox{\textit{minicpm-v-2.6}}, \mbox{\textit{llava-1.6-vicuna-7b}}, and \mbox{\textit{qwen2-vl-7b}}. Each model is configured with a temperature of $0$ and an output length of $2048$.

\subsection{Prompts for \textsc{vlm}s.}

In Table~\ref{tab:data_prompts} and Table~\ref{tab:prompts}, we present the prompts used in \modelname for Vision Language Models. Four distinct prompts are employed: (1) the \textit{Data Processing Prompt}, which employed an answer guided reasoning generation method to prompt \textsc{vlm}s in extracting step-by-step reasoning from YouTube transcripts; (2) the \textit{\macname Prompt}, which is the same as the query in the training data, prompting \textsc{vlm}s to generate a coherent reasoning process to infer the location within an image; (3) the \textit{\micname Prompt}, which generates additional knowledge from image details, and (4) the \textit{\guessname Prompt}, which synthesizes all prior information to make a final prediction.

\begin{table}[htbp]
    \centering
    \small
    \newcolumntype{Y}{>{\arraybackslash}X}
    \begin{tabularx}{\linewidth}{Y}
    \toprule 
        \textit{\textbf{Data Processing Prompt}} \\
        <image> Given an image and the known location details (Country: {country}, Latitude: {lat}, Longitude: {lon}), and an expert's analysis of the location ({transcript}), craft a brief and cohesive reasoning path that deduces this location based on the visual clues present in the image. Begin your reasoning without revealing that you know the exact location, using a tone of exploration and inference. Carefully analyze and link observations of natural features (climate, vegetation, terrain), man-made structures (roads, buildings, signage), and distinct landmarks. Allow these observations to naturally lead you to the correct country, enhancing the accuracy of your deductions. Ensure that while the narrative seems to be guessing, it aligns with the known country, confirming the reliability of your reasoning without stating the specific coordinates. Start the reasoning without any intro, and make sure to make it brief. \\
    \bottomrule
    \end{tabularx}
    \caption{The prompts used in \modelname.}
    \label{tab:data_prompts}
\end{table}

\section{Data.}
\label{apd:data}

In this section, we present the data processing workflows and provide more detailed information on the various types of data used in the system.

\subsection{Data Processing.}

\textbf{YouTubers.} We utilized the scripts of five professional GeoGuessr players' YouTube videos as the starting data for our reasoning generation. We thank these five players for their contributions to knowledge dissemination and promotion of image \geoloc: zi8gzag, GeoWizard, GeoPeter, Geogasm, and RAINBOLT TWO.

\textbf{Data Processing.}
We used the Google Street View\footnote{https://www.google.com/streetview/} API to retrieve images for our dataset. We selected a resolution of 640×640 pixels (the maximum resolution accepted by \textsc{gsv}), a field of view (\textsc{fov}) of 90, and headings of 0, 90, 180, and 270 degrees to obtain four images. Stitching them together produces a complete street view image, providing the same amount of information that a GeoGuessr player would see.

Next, we split the videos for retrieving the transcripts or each round. After a player submits their final guess, the game reveals the distance between their guessed location and the actual coordinates, where the player can choose to either proceed to the next round or end the challenge. We use precise pixel coordinates in conjunction with \textsc{ocr} technology to detect the presence of the ``Next'' or ``End'' buttons and split the videos. We sample frames at a rate of 1/6 per second to ensure no scene is missed. Additionally, we extract the GeoGuessr Score displayed beside the button and collect human players' scores. Next, due to the noise in the data (with many informal language from players), we provide GPT-4o with the correct locations for paraphrasing and generating higher quality and more coherent data.

\begin{table}[htbp]
    \centering
    \small
    \newcolumntype{Y}{>{\arraybackslash}X}
    \begin{tabularx}{\linewidth}{Y}
    \toprule 
        \textit{\textbf{\macname Prompt}} \\
        <image> Given an image, craft a brief and cohesive reasoning path that deduces this location based on the visual clues present in the image. Using a tone of exploration and inference. Carefully analyze and link observations of natural features (climate, vegetation, terrain), man-made structures (roads, buildings, signage), and distinct landmarks. Allow these observations to naturally lead you to the correct country, enhancing the accuracy of your deductions. Start the reasoning without any intro, and make sure to make it brief. \\
        \midrule
        \textit{\textbf{\micname Prompt}} \\
        <image> Analyze the \{item\} images to determine the region with the highest likelihood of finding this type of \{item\}. For each image, provide only the core reasoning in one sentence. Don't say you can't determine, try your best as it's a \geoloc game \\
        \midrule
        \textit{\textbf{\guessname Prompt}} \\
        <image> <information> Using the provided information as a reference, estimate the location depicted in the image with as much accuracy and precision as possible. Generally, you might use the reasoning to roughly locate the coarse-grained location, and use other information to help you decide more precisely. Use your own knowledge as well. Aim to deduce the exact coordinates whenever feasible. Format your response strictly as JSON in the following structure:\{``country'': ``<country\_name>'', ``city'': ``<city\_name>'', ``latitude'': <Latitude Coordinate>, ``longitude'': <Longitude Coordinate>\} Ensure the JSON output is correctly formatted. Provide a well-informed estimate for each value, avoiding any empty fields. Do not include additional information or commentary.\\
    \bottomrule
    \end{tabularx}
    \caption{The prompts used in \modelname.}
    \label{tab:prompts}
\end{table}

\subsection{Data Demonstration.}

In this section, we present examples and key statistics for both \dataname and guidebook datasets.

\begin{table*}[htbp]
\centering
\small
    \newcolumntype{Y}{>{\arraybackslash}X}
    \begin{tabularx}{\linewidth}{Y}
    \toprule 
        \textit{\textbf{Image}} \\
        \includegraphics[width=\linewidth]{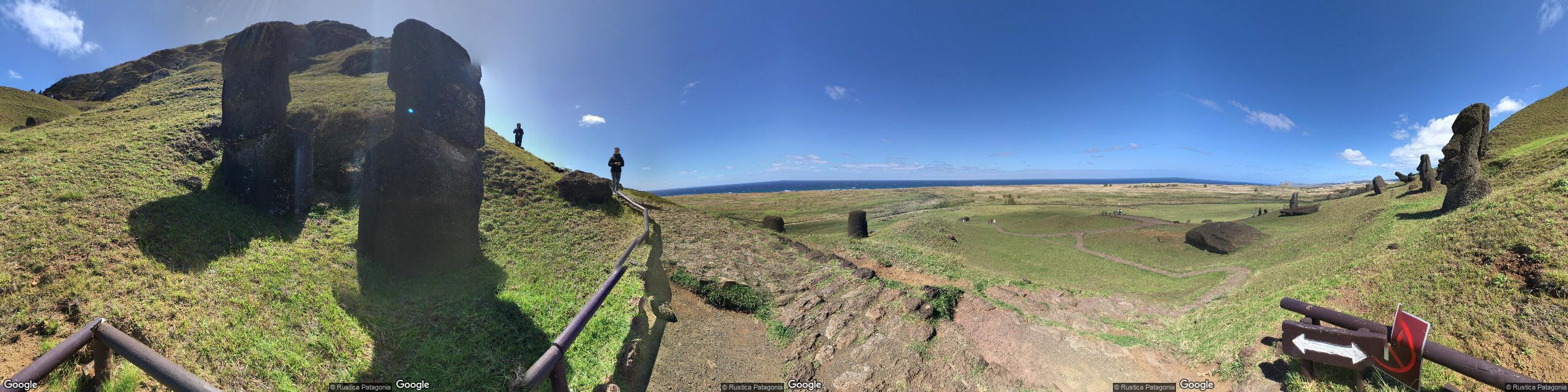} \\
        \midrule
        \textit{\textbf{Location}} \\
        Chile. -27.1265479, -109.2876917 \\
        \midrule
        \textit{\textbf{Reasoning}} \\
        These imposing stone figures are situated on a grassy hillside overlooking the ocean, suggesting a coastal location. The sparse vegetation and the presence of unique megalithic structures indicate a historical and possibly isolated environment. The statues, known as Moai, are characteristic of an island rich in archaeological history. Such a distinct culture is associated with an island famous for these figures, which is set far away from mainland destinations. This points toward the Polynesian island in the southeastern Pacific known for these world-renowned statues, aligning closely with a location associated with Chile. \\
    \toprule 
        \textit{\textbf{Image}} \\
        \includegraphics[width=\linewidth]{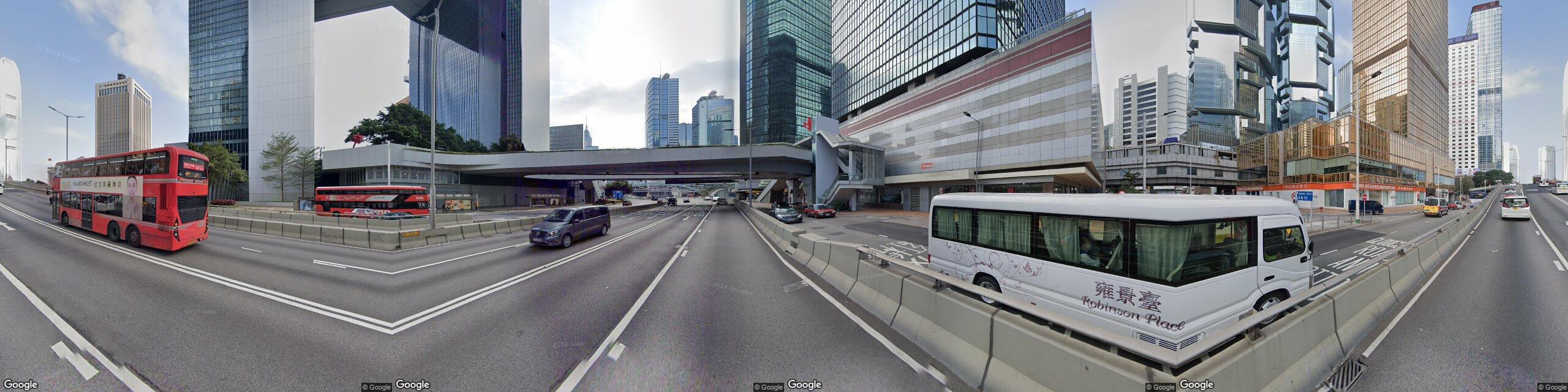} \\
        \midrule
        \textit{\textbf{Location}} \\
        China. 22.27992557066081, 114.1648415981852 \\
        \midrule
        \textit{\textbf{Reasoning}} \\
        Tall, sleek skyscrapers dominate the cityscape, suggesting a prominent urban environment. The distinct yellow license plates on vehicles are a key hint, typically associated with regions influenced by British rule. The street signs feature both English and Chinese characters, which points to a bilingual city. The presence of a distinctive red double-decker bus is highly characteristic of a city with British influence. All these clues, combined with the modern architecture and dense urban vibes, strongly suggest this is Hong Kong, likely within its central business district. \\
    \toprule 
        \textit{\textbf{Image}} \\
        \includegraphics[width=\linewidth]{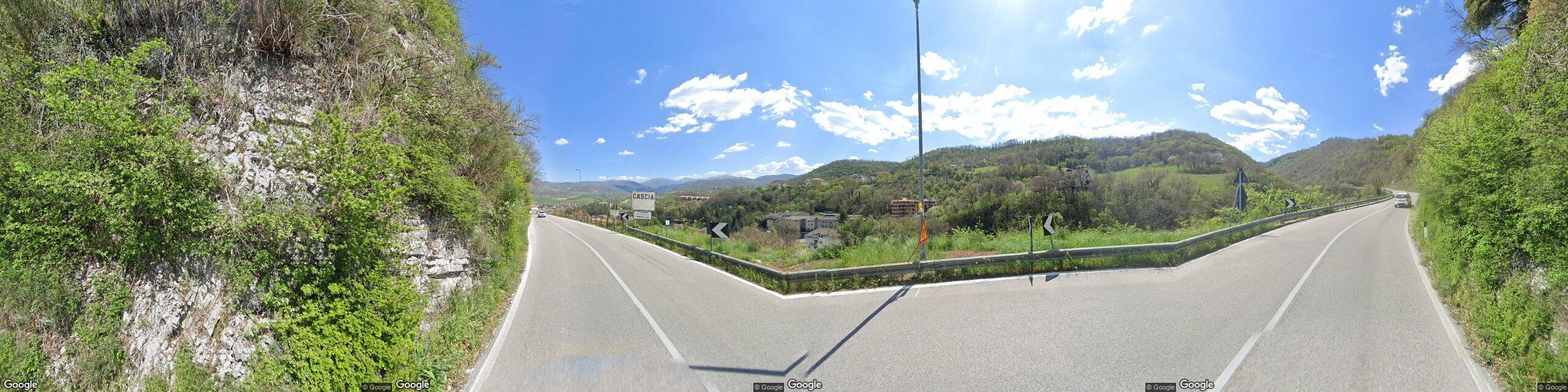} \\
        \midrule
        \textit{\textbf{Location}} \\
        Italy. 42.71658113303754, 13.01051150781922 \\
        \midrule
        \textit{\textbf{Reasoning}} \\
        The landscape features rolling hills and a dense cover of greenery, indicative of a temperate climate typically found in parts of Southern Europe. The road layout with its narrow curve and the types of vehicles suggest a European setting. The presence of Italian-language signage such as Cascia and recognizable Italian road markers, like narrow front plates, strongly points to Italy. The architecture, with its rustic and modest buildings in the distance, complements the rural Italian countryside vibe. Hence, these visual cues collectively affirm the location as Italy, likely in a more central to southern region given the sign for Cascia. \\
    \bottomrule
\end{tabularx}
\caption{
Demonstration of \dataname.
}
\label{tab:geoclues}
\end{table*}

\begin{figure}[htb]
    \centering
    \includegraphics[width=0.8\linewidth]{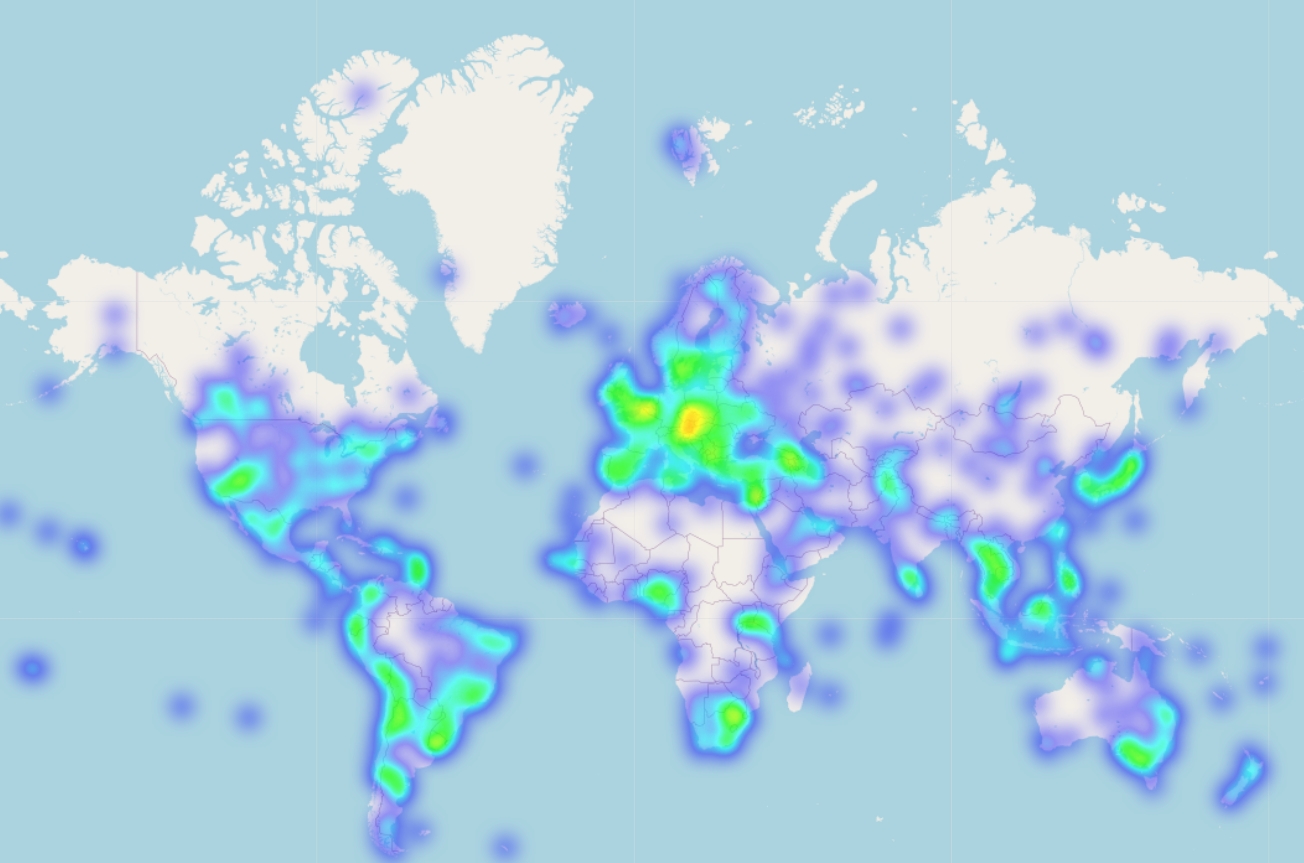} 
    \caption{Location distribution of \dataname, covering a wide range of countries around the world.}
    \label{fig:location_distribution}
\end{figure}

\textbf{\dataname.} 
Each data includes a panoramic image, the corresponding location, and a high-quality reasoning process that shows how geographical and cultural information is used to infer the location (Figure~\ref{tab:geoclues}). To reduce hallucination and bias, the model is not required to generate specific street-level locations or coordinates directly, but carefully analysis about image elements (\textit{e.g.}, climate) that collectively lead to the prediction. \dataname is geographically well-distributed, covering various countries across the globe (Figure~\ref{fig:location_distribution}).

\begin{table}[htbp]
\centering
\small
    \newcolumntype{Y}{>{\arraybackslash}X}
    \begin{tabularx}{\linewidth}{Y}
    \toprule 
        \textit{\textbf{Image}} \\
        \includegraphics[width=\linewidth]{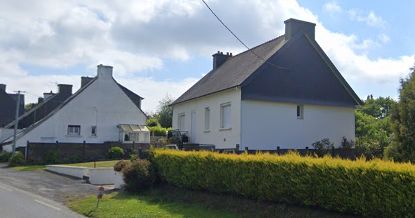} \\
        \midrule
        \textit{\textbf{Text}} \\
        (Toptips) Houses in Brittany, a western region of France, are coloured white with dark roofs. \\
    \toprule 
        \textit{\textbf{Image}} \\
        \includegraphics[width=\linewidth]{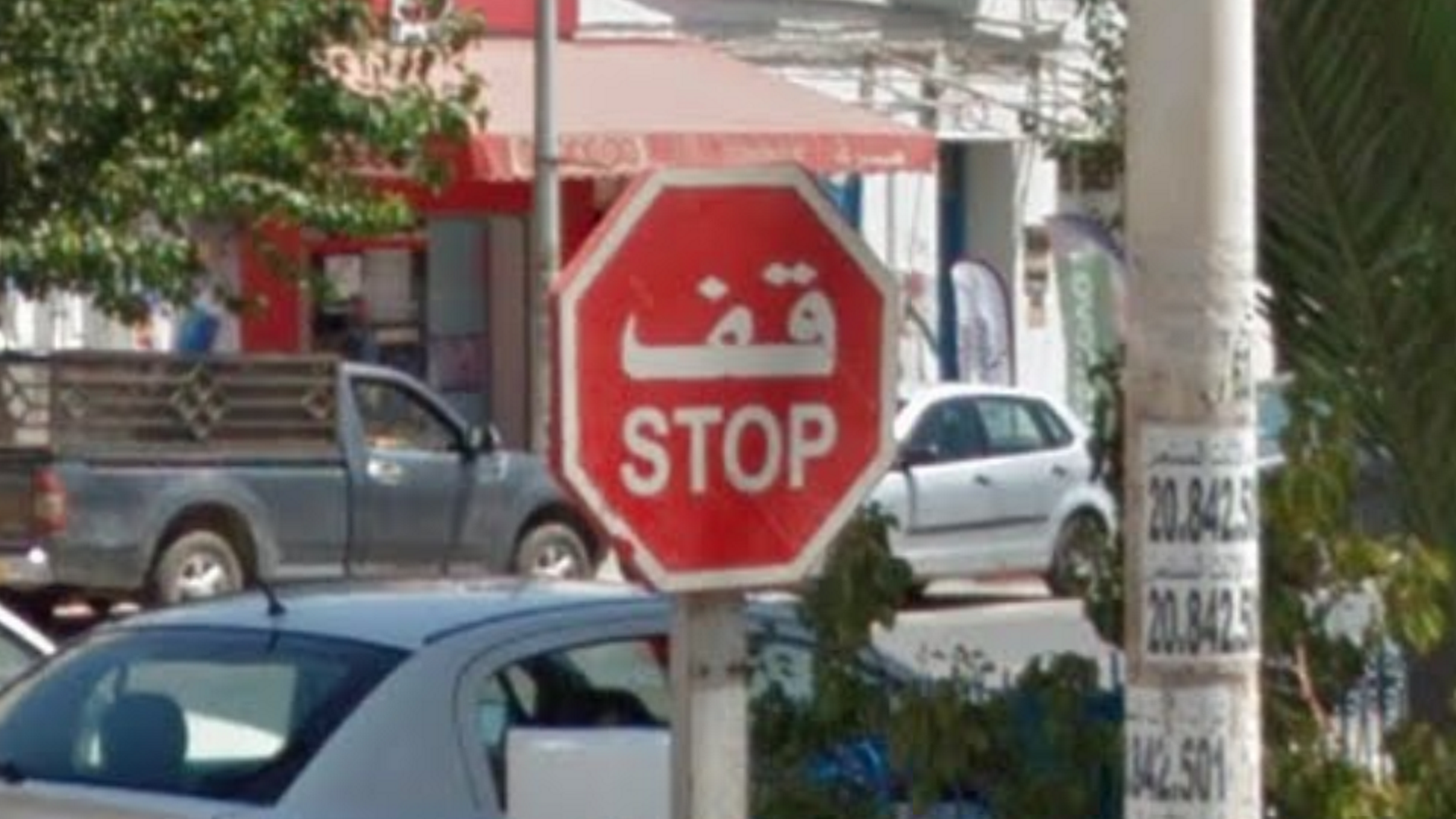} \\
        \midrule
        \textit{\textbf{Text}} \\
        (Plonkit) Tunisia has a fairly unique stop sign with Arabic and Latin script. \\
    \bottomrule
\end{tabularx}
\caption{
Examples of clues in the two guidebooks, Toptips and Plonkit. The data is constructed as image-text pairs in the guidebooks.
}
\label{tab:guidebook}
\end{table}

\textbf{Guidebooks.} As shown in Table~\ref{tab:guidebook}, each guidebook entry consists of an image-text pair, where the text describes the location and features depicted in the image. We utilize \textsc{rag} in \modelname to identify the most similar image and use the accompanying text to support the reasoning process.

\begin{figure*}[t]
    \centering
    \includegraphics[width=0.8\linewidth]{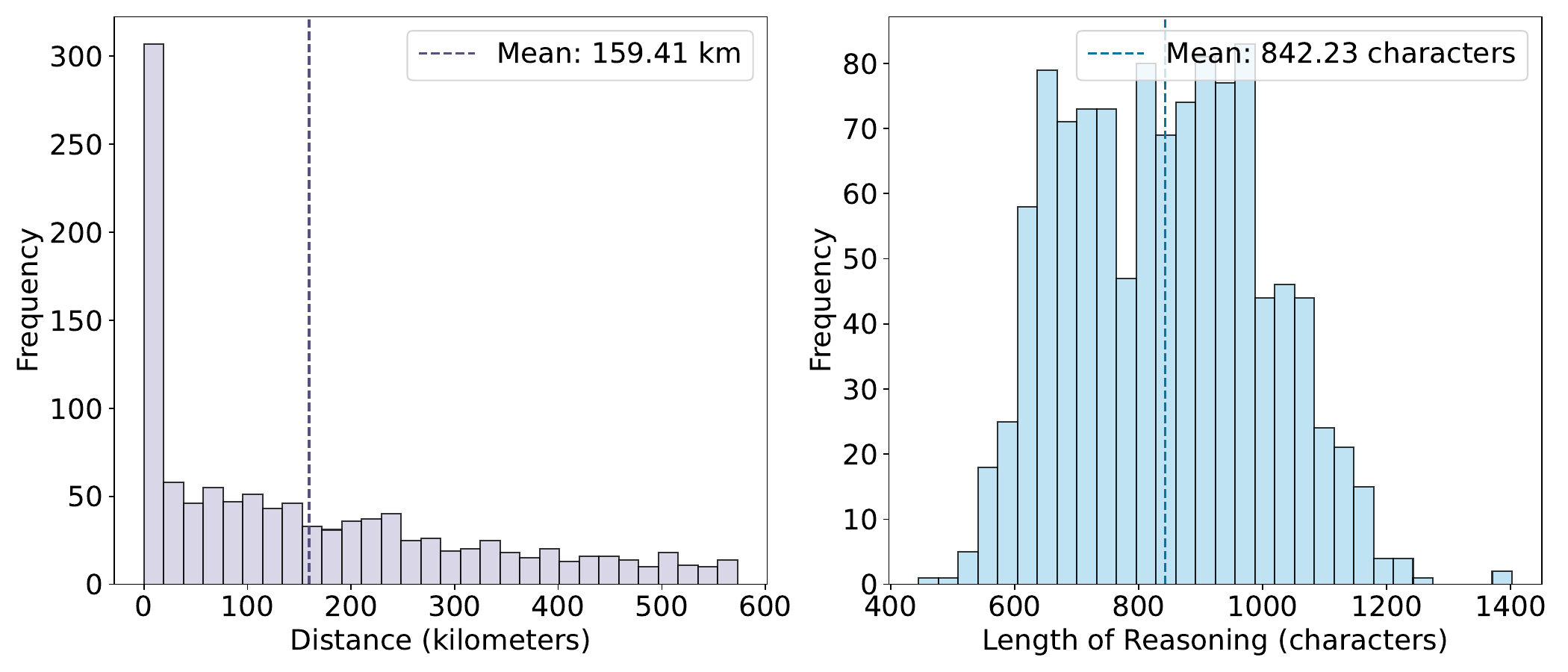} 
    \caption{Distribution of Distance (left) and Length of Reasoning (right) in \dataname.}
    \label{fig:distribution}
\end{figure*}

\textbf{Statistical Results.} We present the statistical results of \dataname in Figure~\ref{fig:distribution}. Most distances are relatively small, indicating that players are often successful in deducing the correct location. The reasoning process averages 842 words in length, demonstrating high quality with extensive details.

\begin{table}[t!]
    \centering
    \small
    \newcolumntype{Y}{>{\arraybackslash}X}
    \begin{tabularx}{\linewidth}{Y}
    \toprule 
        \textit{\textbf{Image}} \\
        \includegraphics[width=\linewidth]{images/israel.jpg} \\
        \midrule
        \textit{\textbf{Location}} \\
        Israel, Ashkelon. 31.66671, 34.59127 \\
        \midrule
        \textit{\textbf{Query}} \\
        <image> Given an image, craft a brief and cohesive reasoning path that deduces this location based on the visual clues present in the image. Using a tone of exploration and inference. Carefully analyze and link observations of natural features (climate, vegetation, terrain), man-made structures (roads, buildings, signage), and distinct landmarks. Allow these observations to naturally lead you to the correct country, enhancing the accuracy of your deductions. Start the reasoning without any intro, and make sure to make it brief. \\
        \midrule
        \textit{\textbf{Response}} \\
        The scene reveals a blend of urban and natural features typical of \textcolor{climate}{\textbf{a Mediterranean climate}}. Notably, the presence of \textcolor{vegetation}{\textbf{palm trees}} suggests a \textcolor{climate}{\textbf{warm, temperate region}}. The architecture of the buildings, characterized by \textcolor{building}{\textbf{flat roofs and a modern, functional style}}, often indicates environments with minimal snowfall. Observing the infrastructure, \textcolor{infrastructure}{\textbf{the neatly paved roads and sidewalks with occasional dust}} align with urban areas in regions with dry summers. The use of earth-toned paint on buildings matches the aesthetic found in Middle Eastern locales. Further exploration of the layout reveals hints of both residential and possibly suburban planning, with open spaces that are common in balanced urban settings. Taking all these elements into consideration, the collective visual attributes and environmental clues point to a location consistent with Israel's landscape and architectural style.\\
    \bottomrule
    \end{tabularx}
    \caption{Fine-tuning data format. The reasoning process leverages visual information in the images to deduce the correct location, such as \textcolor{climate}{\textbf{climate}}, \textcolor{vegetation}{\textbf{vegetation}}, \textcolor{building}{\textbf{building}}, and \textcolor{infrastructure}{\textbf{infrastructure}}.}
    \label{tab:sft_data_full}
\end{table}

\begin{table*}[t]
\centering
\small
\begin{tabular}{l r r r r r r r}
\toprule
\textbf{Model} & \multicolumn{1}{c}{\begin{tabular}[c]{@{}c@{}}\textbf{Continent}\\ $2,500$ km\end{tabular}} & 
\multicolumn{1}{c}{\begin{tabular}[c]{@{}c@{}}\textbf{Country}\\ $750$ km\end{tabular}} & 
\multicolumn{1}{c}{\begin{tabular}[c]{@{}c@{}}\textbf{Region}\\ $200$ km\end{tabular}} & 
\multicolumn{1}{c}{\begin{tabular}[c]{@{}c@{}}\textbf{City}\\ $25$ km\end{tabular}} & 
\multicolumn{1}{c}{\begin{tabular}[c]{@{}c@{}}\textbf{Street}\\ $1$ km\end{tabular}} & 
\multicolumn{1}{c}{\textbf{\begin{tabular}[c]{@{}c@{}}Distance$\downarrow$\end{tabular}}} & 
\multicolumn{1}{c}{\textbf{\begin{tabular}[c]{@{}c@{}}Score$\uparrow$\end{tabular}}}\\
\midrule
$G^3$             & 47.3 & 23.9 &  6.0 &  1.6 &  0.0 & 4,938& 1,451\\
GeoCLIP           & 82.3 & 66.5 & 48.0 & \textbf{32.2} & \textbf{13.0} & 1,762& 3,402\\
StreetCLIP        & 68.2 & 51.2 & 29.6 & 19.0 &  4.2 & 3,161& 2,640\\
\midrule
MiniCPM-V         & 33.2 & 27.8 & 22.4 & 15.9 &  2.3 & 6,624& 1,433\\
LLaVA             & 61.2 & 43.2 & 25.9 & 16.5 &  2.6 & 3,387& 2,338\\
Qwen2-VL          & 75.0 & 65.0 & 48.9 & 29.9 &  5.3 & 2,483 & 3,237\\
\midrule
\textit{\textbf{\modelname}} &&&&&&& \\
~~ - MiniCPM-V    & 68.5& 51.7& 36.5& 23.1 &  3.0 & 3,149 & 2,726 \\
~~ - LLaVA        & 70.4 & 47.8 & 26.8 & 16.7 &  2.8 & 2,851 & 2,592 \\
~~ - Qwen2-VL     & \textbf{84.0} & \textbf{68.3} & \textbf{49.1} &  28.9 & 5.5 & \textbf{1,631} & \textbf{3,482} \\
\bottomrule
\end{tabular}
\caption{
Performance on Im2GPS3k.
}
\label{tab:main_results_im2gps}
\end{table*}

\section{Metrics Definition}
\label{apd:metrics}

\subsection{GeoGuessr Score}

The \textit{GeoGuessr Score} is a metrics introduced in the GeoGuessr game to quantify the accuracy of a guess. It is defined as:

\begin{equation}
    \text{GeoGuessr Score} = 5000 \times e^{-\frac{\text{d}}{1492.7}}
\end{equation}

where $d$ represents the distance between the guessed coordinates and the actual coordinates in kilometers. Therefore, a more accurate guess yields a \textit{GeoGuessr Score} closer to 5,000.

\subsection{Haversine Distance}
We calculate the Haversine Distance of the models with the following formulas: 

\begin{small}
\begin{equation}
\Delta = \sqrt{\sin^2\left(\frac{\delta_\text{lat}}{2}\right) + \cos(\text{lat}_\text{cor}) \cos(\text{lat}_\text{pred}) \sin^2\left(\frac{\delta_\text{lon}}{2}\right)}
\end{equation}

\begin{equation}
d = 2r \cdot \arcsin(\Delta)
\end{equation}
\end{small}

\text{where:}
\begin{itemize}
    \item \( r \) is the Earth’s radius, which we set as 6,371,
    \item \( \delta_\text{lat} \) is the difference in latitude between the true and predicted coordinates,
    \item \( \delta_\text{lon} \) is the difference in longitude between the true and predicted coordinates,
    \item \( \text{lat}_\text{cor} \) and \( \text{lon}_\text{cor} \) are the correct coordinates,
    \item \( \text{lat}_\text{pred} \) and \( \text{lon}_\text{pred} \) are the predicted coordinates.
\end{itemize}

\section{Supplementary Experiments}
\label{apd:supplementary_exps}

In this section, we present supplementary experiments, including results from the experiments on Im2GPS3k, and \micname details.

As shown in Table~\ref{tab:main_results_im2gps}, \modelname outperforms prior models on Im2GPS3k in terms of Average Distance and GeoGuessr Score. However, GeoCLIP achieves better performance at the City and Street level, likely due to its training on coordinates datasets. The ablation results demonstrated in Table~\ref{tab:ablation_results_im2gps} are consistent with those in Table~\ref{tab:ablation_results}.

\begin{table}[htbp]
\centering
\small
\begin{tabular}{l r r r}
\toprule
\textbf{Model} & \textbf{Country} & \textbf{City} & \textbf{Street} \\
\midrule
 \textbf{\modelname}~(MiniCPM-V) & \textbf{51.7} &  \textbf{23.1} &  \textbf{3.0} \\
~~ - w/o training             &- 1.6 & - 1.8 & - 0.1 \\
 ~~ - w/o \macname                &- 10.6 & - 3.5 & - 0.2 \\
~~ - w/o \micname                & - 0.3 & - 0.2 & - 0.0 \\
 ~~ - MiniCPM-V                &- 23.9 & - 7.2 & - 0.7 \\
\midrule
 \textbf{\modelname}~(LLaVA) & 47.8 & \textbf{16.7} & \textbf{ 2.8} \\
~~ - w/o training             &- 15.3 & - 4.7 & - 0.8 \\
 ~~ - w/o \macname                & - 8.1 & - 1.3 & - 0.1 \\
~~ - w/o \micname                & \textbf{+ 0.1} & - 0.2 & - 0.2 \\
 ~~ - LLaVA                    &- 4.5 & - 0.2 & - 0.1 \\
\midrule
 \textbf{\modelname}~(Qwen2-VL) & 68.3 &  28.9 &  \textbf{5.5} \\
~~ - w/o training             & - 4.3 & - 1.2 & - 0.3 \\
 ~~ - w/o \macname                & - 2.9 & + 0.5 & - 0.1 \\
~~ - w/o \micname                & \textbf{+ 0.1} & - 0.0 & - 0.2 \\
 ~~ - Qwen2-VL                 & - 3.3 & \textbf{+ 1.0} & - 0.2 \\
\bottomrule
\end{tabular}
\caption{
Ablation results on Im2GPS3k.
}
\label{tab:ablation_results_im2gps}
\end{table}

We also analyze the usage of each tool across the datasets and the number of grounding images. This analysis illustrates how frequently \modelname leverages each tool and image detail to deduce locations. As shown in Table~\ref{tab:grounding_distribution} and Table~\ref{tab:tool_usage}, houses are the most frequently identified items in the testing dataset, as images often contain multiple houses. In contrast, signs, though less common, play a critical role by generating queries for \textsc{osm}. The distribution of items directly influences the frequency of tool usage for knowledge retrieval.

\begin{table}[htbp]
\centering
\small
\begin{tabular}{l r r r}
\toprule
\textbf{Dataset} & \textbf{house} & \multicolumn{1}{c}{\begin{tabular}[c]{@{}c@{}}\textbf{road} \\ \textbf{sign}\end{tabular}} & \multicolumn{1}{c}{\begin{tabular}[c]{@{}c@{}}\textbf{building} \\ \textbf{sign}\end{tabular}} \\
\midrule
GWS5k           & 3,451 &  20 & 104 \\
Im2GPS3k        & 465   &  52 & 24 \\
\bottomrule
\end{tabular}
\caption{
The frequency of how each item is grounded.
}
\label{tab:grounding_distribution}
\end{table}

\begin{table}[htbp]
\centering
\small
\begin{tabular}{l c r r r}
\toprule
\textbf{Dataset} & $N$ & \textbf{RAG} & \textbf{MAP} & \textbf{\textsc{vlm}} \\
\midrule
GWS5k            & 5,000 & 128 & 70 & 1,978 \\
Im2GPS3k         & 2,997 &  213 & 21 & 493 \\
\bottomrule
\end{tabular}
\caption{
The usage of each tool in each dataset, where $N$ denotes the size of the dataset.
}
\label{tab:tool_usage}
\end{table}

\end{document}